\documentclass[10pt,twocolumn,letterpaper]{article}

\usepackage{cvpr}
\usepackage{times}
\usepackage{epsfig}
\usepackage{graphicx}
\usepackage{amsmath}
\usepackage{amssymb}
\usepackage{enumitem}
\usepackage{subcaption}
\usepackage{fancyhdr}
\usepackage[ruled,vlined]{algorithm2e}
\include{pythonlisting}
\usepackage{booktabs}
\usepackage{adjustbox}
\usepackage{multirow}
\newcommand{\punt}[1]{}

\newcommand{\reals}{\mathbb{R}}

\newcommand{\bq}{\begin{equation}}
\newcommand{\eq}{\end{equation}}
\newcommand{\ba}{\begin{eqnarray}}
\newcommand{\ea}{\end{eqnarray}}

\newcommand{\remove}[1]{}

\makeatletter
\DeclareRobustCommand\onedot{\futurelet\@let@token\@onedot}
\def\@onedot{\ifx\@let@token.\else.\null\fi\xspace}
 
\def\ie{\emph{i.e}\onedot} 

\makeatother

\usepackage[pagebackref=true,breaklinks=true,letterpaper=true,colorlinks,bookmarks=false]{hyperref}

 \cvprfinalcopy 

\def\cvprPaperID{****} 

\ifcvprfinal\fi
\begin{document}

\title{The Riemannian Geometry of Deep Generative Models}

\author{Hang Shao\\
University of Utah\\
Salt Lake City, UT\\
{\tt\small hang.shao@utah.edu}
\and
Abhishek Kumar\\
IBM Research AI\\
Yorktown Heights, NY\\
{\tt\small abhishk@us.ibm.com}
\and
P. Thomas Fletcher\\
University of Utah\\
Salt Lake City, UT\\
{\tt\small fletcher@sci.utah.edu}
}

\maketitle

\begin{abstract}
Deep generative models learn a mapping from a low-dimensional latent space to a high-dimensional data space. Under certain regularity conditions, these models parameterize nonlinear manifolds in the data space. In this paper, we investigate the Riemannian geometry of these generated manifolds. First, we develop efficient algorithms for computing geodesic curves, which provide an intrinsic notion of distance between points on the manifold. Second, we develop an algorithm for parallel translation of a tangent vector along a path on the manifold. We show how parallel translation can  be used to generate analogies, i.e., to transport a change in one data point into a semantically similar change of another data point. Our experiments on real image data show that the manifolds learned by deep generative models, while nonlinear, are surprisingly close to zero curvature. The practical implication is that linear paths in the latent space closely approximate geodesics on the generated manifold. However, further investigation into this phenomenon is warranted, to identify if there are other architectures or datasets where curvature plays a more prominent role. We believe that exploring the Riemannian geometry of deep generative models, using the tools developed in this paper, will be an important step in understanding the high-dimensional, nonlinear spaces these models learn.
\end{abstract}

\section{Introduction}              
Learning from unlabeled raw sensory observations, which are often high-dimensional, is a problem of significant importance in machine learning. An influential notion in this line of research is the \emph{manifold hypothesis}, which states that these high-dimensional observations are concentrated around a manifold of much lower dimensionality \cite{fefferman2016testing,roweis2000nonlinear,saul2003think,tenenbaum1998mapping,tenenbaum2000global}. Indeed, the manifold hypothesis has been the basis of much of the prior work on the problems of unsupervised and semi-supervised learning \cite{tenenbaum1998mapping,roweis2000nonlinear,tenenbaum2000global,saul2003think,vincent2003manifold,belkin2004semi,cayton2005algorithms,belkin2006manifold,weinberger2006unsupervised,rifai2011manifold,kumar2017improved}. 


These problem areas have witnessed a surge in activity following the recent success of deep generative models in modeling the observed data with higher fidelity than was earlier possible. This is particularly true for visual observations, where deep generative models such as variational autoencoders (VAEs) \cite{kingma2013auto,rezende2014stochastic}, generative adversarial networks (GANs) \cite{goodfellow2014generative},  PixelCNN \cite{oord2016pixel}, and their variants \cite{radford2015unsupervised,zhao2017towards,berthelot2017began,karras2017progressive} have been shown to generate good quality images. All of these models involve learning a mapping (termed as \emph{generator} or \emph{decoder}) from a lower-dimensional latent space to the high-dimensional space of observed data. This allows for generating novel data samples by ancestral sampling, which is seeded by samples in the latent space. 

As the learned generator in these models is able to generate high-fidelity data samples, the generator mapping can be argued to approximate the data manifold reasonably well. This has been explored in the context of semi-supervised learning to obtain smooth invariances for classification via estimating the tangent directions to the data manifold as learned by the generator \cite{kumar2017improved,rifai2011manifold}. However, the {\it metric} properties of these generated manifolds still remain unexplored. 

In this work, we investigate the Riemannian geometry of the manifolds learned by these deep generative models. Our contributions are summarized as follows:
\begin{itemize}[noitemsep,topsep=0pt,parsep=0pt,partopsep=0pt,leftmargin=*]
\item We propose an algorithm for computing {\bf geodesic paths} between points on the generated manifold. This can be used to interpolate between two generated data points on the manifold using the least amount of change necessary, while enforcing that the points along the path remain on the manifold. The arclength of a minimal geodesic path is a distance metric between points on the manifold, and is a natural way to measure the similarity between two data points. While the continuous geodesic equation requires expensive second derivatives and matrix inversions, we formulate an efficient numerical strategy for computing discretized geodesic curves that avoids these computations. In addition to point-to-point geodesics paths, we show how to ``shoot'' a geodesic from an initial starting position and initial velocity (tangent vector).
\item Next, we develop an algorithm for {\bf parallel translation} of tangent vectors along a path on the generated manifold. Parallel translation moves a tangent vector continuously along a path using the minimal amount of change needed to keep it tangent to the manifold. This operation provides a means for computing analogies, i.e., taking the change between points $a$ to $b$ on the manifold (represented as a geodesic segment) and applying that change to a third point $c$.
\item In our experiments, we show how the above tools can be used to {\bf explore the Riemannian geometry} of the manifolds learned by deep generative models, and in particular, to investigate the {\bf curvature} of these manifolds. We demonstrate, at least for the VAE architecture used in our experiments, that the generated manifolds learned from real images used in our experiments (CelebA \cite{liu2015deep}, SVHN \cite{netzer2011reading}) have surprisingly little curvature. As a result, straight lines in the latent space map via the generator to curves on the manifold that are quite similar to geodesics. This may help explain why the latent coordinates, and interpolations between them, tend to give plausible changes in the generated images. Geodesic curves are in a sense the smoothest possible transitions, as they move at constant speed and minimize the amount of distance needed to travel from one point to another. Our conclusion is that latent coordinates that approximate geodesics is a desirable property to have, and this should be checked by interrogating the Riemannian geometry of a trained deep generative model.
\end{itemize}

\section{Deep Generative Models as Manifolds}

\begin{figure}
\includegraphics[width=\linewidth]{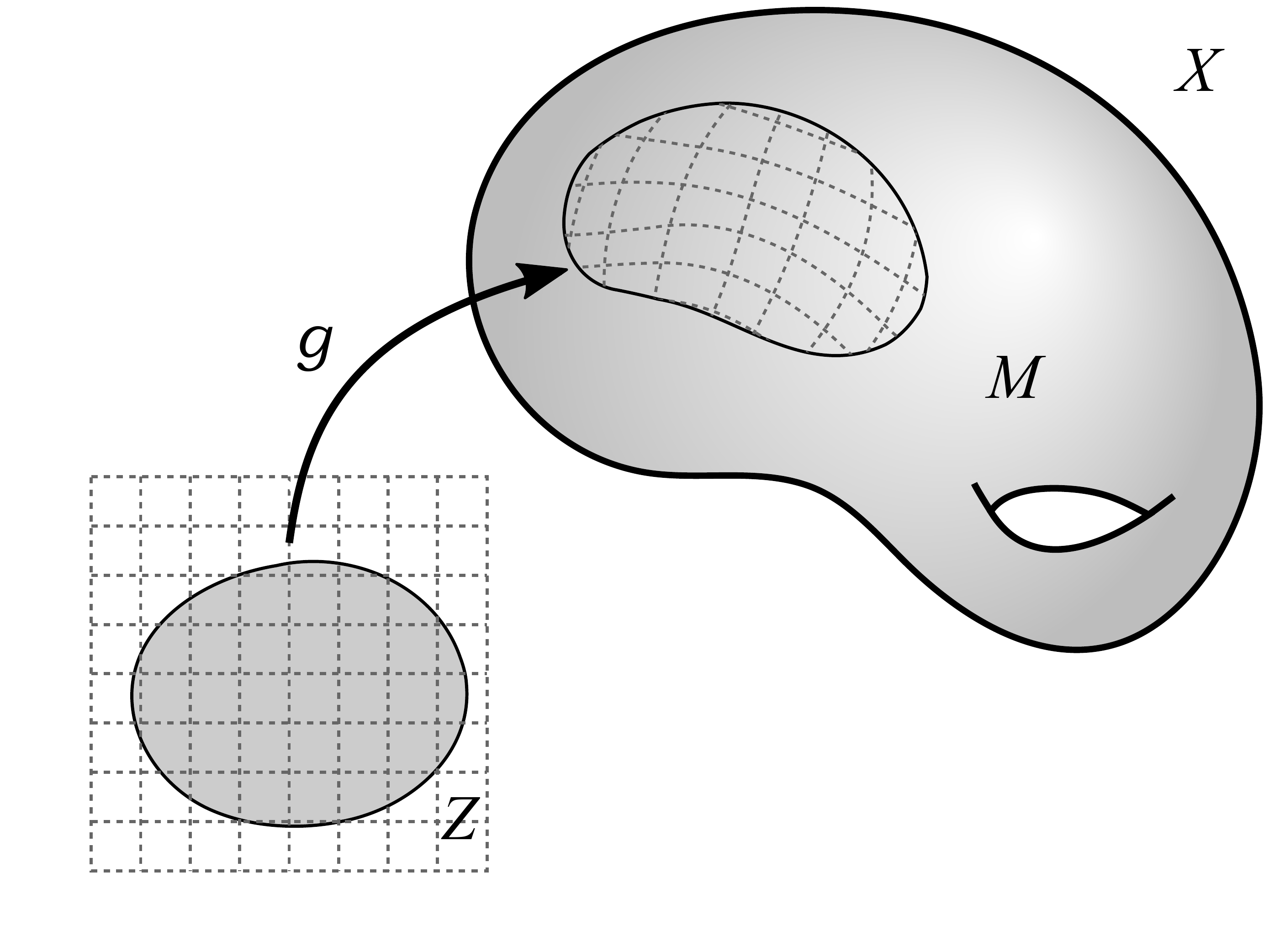}
\vspace{-10mm}
\caption{Depiction of a generative model as a mapping from low-dimensional latent coordinates onto an immersed manifold in data space.}
\label{fig:manifold}
\vspace{-3mm}
\end{figure}

In this section, we illustrate the connection between deep generative models and manifolds. A deep generative model represents a mapping, $g : Z \rightarrow X$, from some low-dimensional latent space $Z \subseteq \mathbb{R}^d$ to a high-dimensional data space $X \subseteq \mathbb{R}^D$ (typically, $d \ll D$). Under certain conditions (described precisely below), the image of $g$ is a smooth manifold, $M \subset X$. As depicted in Figure~\ref{fig:manifold}, $g$ maps linear coordinates in $Z$ onto curvilinear coordinates on $M$.

We will construct $g$ as the composition of multiple layers, i.e., $g = g^{(1)} \circ g^{(2)} \circ \cdots \circ g^{(l)}$, where we use superscripts to denote the layer index. Each layer $g^{(l)}$ is an affine mapping, followed by a nonlinear activation function:
$$g_k^{(l)}\left(y^{(l)}\right) = \phi\left(W_k^{(l)} y^{(l)} + b^{(l)}\right).$$
Here we have used subscripts, $g_k^{(l)}$, to denote the $k$th component of the output, and $W_k^{(l)}$, to denote the $k$th row of the weight matrix, $W^{(l)}$. 

The image of $g$ is a smooth (i.e., $C^\infty$), $d$-dimensional, immersed manifold if for every point $z \in Z$, the Jacobian of $g$ at $z$, $J_g(z)$, has rank $d$. As a straightforward application of the chain rule, this will be true when the following conditions are met:
\begin{enumerate}[noitemsep]
\item The activation function, $\phi$, is a smooth, monotonic function.
\item Each weight matrix $W^{(l)}$ has maximal rank.
\end{enumerate}

Note that condition 1 can be enforced during the modeling phase, by selecting an appropriate activation function. Condition 2 must be checked after training. Also, note that condition 2 is sufficient but not necessary: we could potentially have less-than-maximal rank weight matrices in the middle layers, as long as the final rank of the Jacobian is $d$. However, checking this more general condition would require checking the Jacobian is rank-$d$ at every possible input $z$, which is not feasible. Finally, we emphasize that $M$ is only guaranteed to be an {\it immersed} manifold. This means that it is locally diffeomorphic to $d$-dimensional Euclidean space, but globally it may have self intersections.

The Jacobian matrix of $g$ provides a way to map tangent vectors in the latent space to tangent vectors on the manifold. At any point $z \in Z$, the Jacobian matrix $J_g(z)$ is a linear mapping from $T_z Z$, the tangent space of $Z$ at $z$, to $T_{g(z)} M$, the tangent space of $M$ at $g(z)$. In practice, $J_g(z)$ is computed as the $d \times D$ partial derivative matrix of $g$ via backpropagation. A {\it Riemannian metric} provides an inner product structure between tangent vectors in each tangent space $T_x M$. We will use the induced metric from the ambient data space $X$. In other words, thinking of two vectors $u, v \in T_x M$ as living in a linear subspace of $X$, we can use the Euclidean dot product of $X$ to compute the Riemannian metric $\langle u, v \rangle$.

Intuitively speaking, the curvature of a Riemannian manifold measures the extent to which the metric deviates from being Euclidean. For a precise mathematical explanation of curvature, refer to standard texts in Riemannian geometry, e.g., \cite{doCarmo}. We emphasize an important distinction: just because a manifold is {\em flat}, i.e., has zero curvature, does not mean that it isn't nonlinear. For example, take a sheet of paper and draw a straight line on it. Now bend the sheet of paper into any shape without creasing it. This surface is metrically equivalent to 2D Euclidean space: the straight line you drew is now a geodesic curve with the same arc length. In other words, the surface has zero curvature (this is the Gaussian curvature in the case of a 2D surface). For example, rolling the paper into the famous ``swiss roll'' results in a surface that is highly nonlinear, but nonetheless has zero curvature.

\section{Riemannian Geometry Computations}
In this section we develop three algorithms for Riemannian computations on a manifold represented by a deep generative network $g$. These are {\bf geodesic interpolation} between two points on the manifold, {\bf parallel translation} of a tangent vector along a path on the manifold, and {\bf geodesic shooting} from an initial point and velocity on the manifold. We begin with a general discussion of the geodesic equation on a Riemannian manifold.

We will consider all objects (tangent vectors, curves, the Riemannian metric) to be defined in the coordinate space $Z$. However, we point out that all of these objects each have a corresponding unique counterpart on the manifold, $M$, through the mapping $g$ (or it's derivative mapping). We represent the Riemannian metric as a symmetric, positive definite matrix field, $G(z)$, defined at each point of the latent coordinate space, $z \in Z$. It is given by the formula:
$$G(z) = J_g(z)^T J_g(z).$$
Given two tangent vectors $u, v \in T_z Z$ in coordinates, their inner product is $\langle u, v\rangle = u^T G(z) v$.

Now, consider a smooth curve $\gamma : [a, b] \rightarrow Z$. Again, this corresponds to a curve on the manifold, $g \circ \gamma(t) \in M$. The arc length of $\gamma$ is defined as 
\begin{align}
L(\gamma) = \int_a^b \sqrt{\dot{\gamma}(t)^T G_{\gamma(t)} \dot{\gamma}(t)} dt.
\end{align}
A geodesic curve locally minimizes the arc length, although this is done through minimizing a slightly different energy functional:
\begin{align}
\label{Eq:energy}
E(\gamma) = \frac{1}{2}\int_a^b \dot{\gamma}(t)^T G_{\gamma(t)} \dot{\gamma}(t) dt.
\end{align}
Minimizing this energy leads to geodesic curves, which also locally minimize the arc length, but in addition have constant speed parameterizations.

Taking a variation of the geodesic energy functional \eqref{Eq:energy} results in the Euler-Lagrange equation for a geodesic:
\begin{align}
\frac{d^2 \gamma^{i}}{dt^2} = -\Gamma^{i}_{jk}\frac{d\gamma^{j}}{dt}\frac{d\gamma^{k}}{dt},
\label{eq:geodesic}
\end{align}
where $\Gamma^{i}_{jk}$ are the Christoffel symbols of the metric $G$. These are defined as
$$\Gamma^{i}_{jk} = \frac{1}{2}G^{il}\left(\frac{\partial G_{lj}}{\partial x^{k}}+\frac{\partial G_{lk}}{\partial x^{j}}-\frac{\partial G_{jk}}{\partial x^{l}}\right),$$
where $G^{il}$ is inverse of $G_{il}$. Geodesic paths can then be computed using a numerical integration of the ordinary differential equation \eqref{eq:geodesic}. However, notice that computation of the Christoffel symbols requires taking derivatives of $G$ (which involves second derivatives of the generator, $g$) and also a matrix inverse of $G$. As we show in the next subsection, these expensive calculations can be avoided if we start from a discrete counterpart to the geodesic energy \eqref{Eq:energy}.

\subsection{Efficient Discrete Geodesic Computation}
We begin with a discretized curve as a sequence of coordinates $z_0, z_1, \ldots, z_T \in Z$. We think of this as approximating a continuous curve, $\gamma : [0, 1] \rightarrow Z$. Thus, with $T$ time steps, we have a discrete time interval of $\delta t = 1 / T$. This also corresponds to a discrete curve on the manifold $M$ as $g(z_i)$. Using forward finite differences, we get the approximate velocity of the curve at $g(z_i)$ as $u_i = (g(z_{i+1}) - g(z_i)) / \delta t$. Now the discrete analog \eqref{Eq:energy} gives us the energy of this curve:
\begin{align}
E_{z_i} = \frac{1}{2}\sum_{i=0}^T \frac{1}{\delta t} \|g(z_{i+1}) - g(z_{i})\|^2.
\end{align}

Fixing the endpoints, $z_0$ and $z_T$, as our target start and end points of the geodesic path, we will minimize this discrete geodesic energy by taking a gradient descent in the remaining points on the curve, $z_1, \ldots, z_{T-1}$. The gradient with respect to $z_i$ is
\begin{align}
\nabla_{z_i} E  = -\frac{1}{\delta t} J_g^T(z_i) \left(g(z_{i+1}) - 2g(z_i) + g(z_{i-1}) \right).
\label{eq:grad}
\end{align}
Notice that the gradient is a finite-difference second derivative in the $X$ space, followed by a Jacobian of $g$ coming from the chain rule. The second finite difference in $X$ space may have a component normal to the tangent space $T_{g(z_i)} M$. However, the $J_g^T$ will project out this normal component and map the gradient in $X$ to a gradient in $T_{z_i} Z$. Finally, geodesic path finding proceeds by optimizing the curve coordinates $z_1, \ldots, z_{T-1}$, using gradient descent with the gradient in \eqref{eq:grad}.

\begin{algorithm}[h]
\KwIn{Two points,$z_0, z_T\in Z$\\ $\alpha \in \mathbb{R_+}$ is gradient descent step size}
\KwOut{Discrete geodesic path ,$z_0, z_1, \ldots, z_T \in Z$}
    \caption{{\bf Geodesic Path} \label{alg:geodesic}}
Initialize $z_i$ as linear interpolation between $z_0$ and $z_T$
\While{$\sum_i \|\nabla_{z_i} E\|^2 > \epsilon$}{
\For{$i \in \{1, \ldots, T - 1\}$}{
 Compute the modified gradient $\eta_i$ using \eqref{eq:mod-grad}\\
 $z_i \leftarrow z_i - \alpha \eta_i$\\
}
}
\end{algorithm}

While this gradient descent algorithm for computing discretized geodesics avoids the expensive Christoffel symbol calculations, it does still require computation of the Jacobian of the generator, $g$. For deep generative models, this Jacobian can be expensive. However, we can make an additional speed up for models with a corresponding encoder function, i.e., a mapping $h : X \rightarrow Z$, such that $h(g(z)) = z$. For such models, e.g., VAEs, the encoder Jacobian is significantly faster. Now imagine moving our discrete curve points, $g(z_i)$, in the negative gradient direction along $M$. Mapping this direction down into $Z$ via the Jacobian of $h$, $J_h(g(z_i))$, will produce an equivalent direction in $Z$ coordinates. This results in the following modified gradient, which replaces $J_g^T$ with the faster-to-compute $J_h$:
\begin{align}
\eta_i  = -\frac{1}{\delta t} J_h(g(z_i)) \left(g(z_{i+1}) - 2g(z_i) + g(z_{i-1}) \right).
\label{eq:mod-grad}
\end{align}

Although this modified gradient is no longer the gradient of the discrete curve energy, it does move the $g(z_i)$ in the same initial direction. Also, descent in this modified gradient direction has the same fixed point as gradient descent. The final geodesic path algorithm is given in Algorithm \ref{alg:geodesic}.



\subsection{Parallel Translation}
Given a geodesic path from a point $a \in M$ to a point $b \in M$, we can transfer the change from $a \rightarrow b$ into a change of a third point $c \in M$. This type of ``analogy'' is performed in three steps: (1) compute the initial velocity to the geodesic from $a$ to $b$, (2) parallel translate this velocity along the geodesic from $a$ to $c$, and (3) use this velocity at $c$ to shoot a geodesic segment. In Euclidean space, these operations would be (1) take the difference $v = b - a$, (2) consider $v$ as a vector based at $c$, and (3) shoot the geodesic (straight line) by adding $c + v$. Parallel translation for non-flat manifolds moves a tangent vector along the manifold with as little change as possible, while still enforcing the vector stay tangent. This operation preserves the inner product between tangent vectors, and as such, preserves the length of a translated tangent vector. As a concrete example, imagine the 2D sphere with a tangent vector at the north pole. Now rotate the sphere and tangent vector with it. This is parallel translation along the path swept out by the rotation.

Now, assume that we already have a discrete path $z_0, \ldots z_T \in Z$ in coordinates and a tangent vector in $u_0 \in T_{g(z_0)} M$ at the initial point on the manifold. A small step of parallel translation is approximately equivalent to Euclidean translation of the vector $u_0$ from $g(z_0)$ to $g(z_1)$. However, the vector at this new position will be slightly out of the tangent space. This can be corrected by applying the minimal rotation to bring this vector into the tangent space. Note that we can do this using the singular value decomposition (SVD) of the Jacobian $J_g$. The left singular vectors $U$ give an orthonormal basis for the tangent space. Rotation onto this basis is equivalent to a projection (multiplication by $U^T U$) followed by a rescaling of the vector back to it's original length. Repeating this for process for each time step along the curve gives our parallel translation routine, summarized in Algorithm \ref{alg:parallel}.

\begin{algorithm}
\caption{\bf Parallel Translation}
\label{alg:parallel}
\KwIn{Discrete path: $z_0, z_1, \ldots, z_T \in Z$, and
tangent vector: $v_0 \in T_{z_0}Z$}
\KwOut{Tangent vector $v_T \in T_{z_T}Z$}
Initialize: $u_0 = J_g(z_0) v_0$\\
\For{$i = 0, \ldots, T-1$}{
$x_i = g(z_i)$\\
Compute SVD: $J_g(z_{i+1}) = U \Sigma V^T$\\
$u_{i+1} = U^T U u_i$\\
$u_{i+1} = \frac{\|u_i\|}{\|u_{i+1}\|} u_{i+1}$\\
}
$v_T = J_h(x_T) u_T$
\end{algorithm}

\subsection{Geodesic Shooting}
Given a starting point $x_0 \in M$ and a starting velocity $u_0 \in T_{x_0} M$, there is a unique geodesic $\gamma(t) \in M$, with these initial conditions $\gamma(0) = x_0$ and $\dot{\gamma}(0) = u_0$. (Technically, such a geodesic is only guaranteed to exist for some finite time.) In Euclidean space, this intuitively says that given a starting point and velocity, there is only one straight line with those initial conditions.

To compute geodesic shooting, that is, a geodesic path from initial conditions, we will use the connection between the geodesic equation and parallel translation from the previous subsection. The geodesic equation says that the velocity of a geodesic moves by parallel translation along the geodesic. Therefore, we can compute a discrete geodesic step by taking a small step in the current velocity direction, followed by updating the velocity to this new point by parallel translation. This process is detailed in Algorithm \ref{alg:shoot}.


\begin{algorithm}[h]
\caption{\bf Geodesic Shooting}
\label{alg:shoot}
\KwIn{Initial point $z_0 \in Z$, and initial velocity $u_0 \in T_{g(z_0)} M$}
\KwOut{Final point on geodesic segment: $z_T \in Z$}
Discrete timestep: $\delta t = \frac{1}{T}$\\
\For{$i = 0, \ldots, T-1$}{
$x_{i+1} = x_{i} + \delta t u_i$\\
$z_{i+1} = h(x_{i+1})$\\
$x_{i+1} = g(z_{i+1})$\\
Compute SVD: $J_g(z_{i+1}) = U \Sigma V^T$\\
$u_{i+1} = U^T U u_i$\\
$u_{i+1} = \frac{\|u_i\|}{\|u_{i+1}\|} u_{i+1}$\\
}
\end{algorithm}

\section{Experiments}
\begin{figure*}[t]
    \begin{center}
  \includegraphics[width=0.33\textwidth]{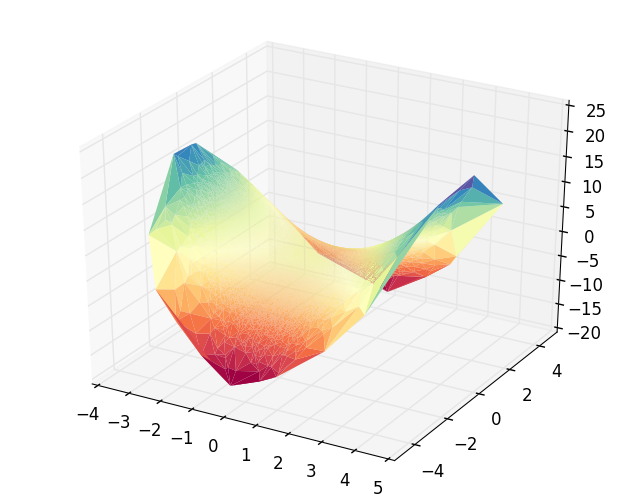}\includegraphics[width=0.33\textwidth]{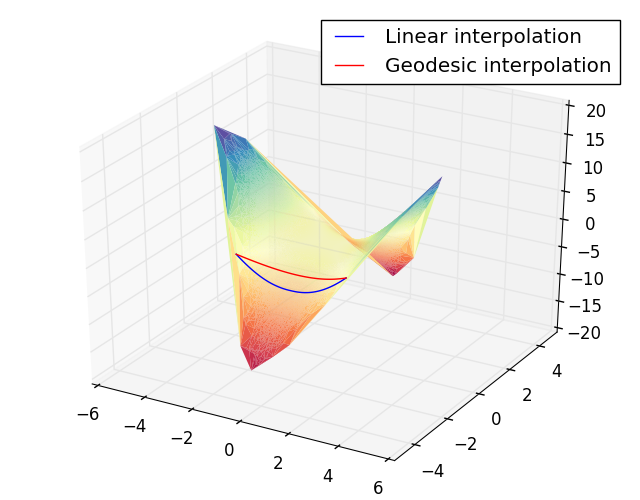}\includegraphics[width=0.25\textwidth]{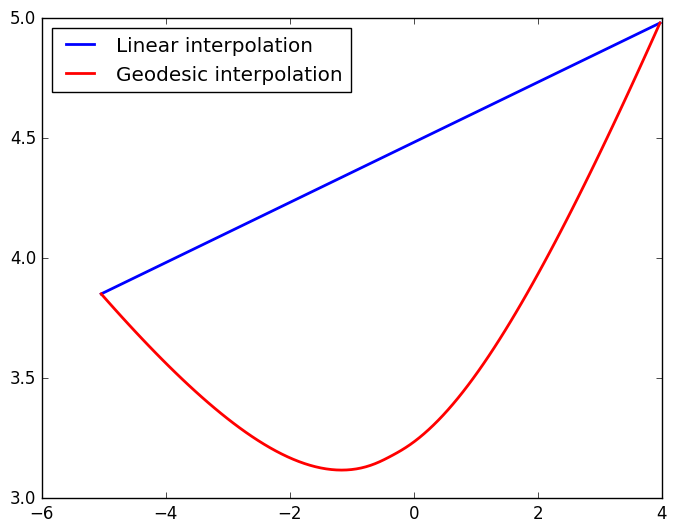}
\end{center}
\vspace{-5mm}
\caption{Comparing linear interpolation with geodesic interpolation (Algorithm \ref{alg:geodesic}) for a pair of points on the manifold ${M}$ induced by the generator of the VAE, which is trained on the data from a hyperbolic paraboloid.  \emph{Left:} True surface of the hyperbolic paraboloid, \emph{Middle:} Surface of $M$ (range of the VAE's learned generator mapping, $g:Z\to X$) overlaid with the curves for linear and geodesic interpolation, \emph{Right:} Linear and geodesic interpolation curves in $Z$.}
\label{fig:synthgeo}
\end{figure*}

In this section, we conduct an extensive empirical study of the proposed algorithms for various Riemannian geometry computations in the context of deep generative models.
We work with variational autoencoder (VAE) \cite{kingma2013auto,rezende2014stochastic} as our generative model of choice, however, the proposed algorithms are equally applicable to other popular generative models, such as generative adversarial network \cite{goodfellow2014generative} and PixelCNN \cite{oord2016pixel}. 

\begin{table}[t]
\centering
\begin{tabular}{c } 
\toprule
VAE Encoder architecture \\
\midrule
Conv $4\times 4\times 32$ (stride 2), Batch norm, ELU   \\
Conv $4\times 4\times 32$ (stride 2), Batch norm, ELU  \\
Conv $4\times 4\times 64$ (stride 2), Batch norm, ELU  \\
Conv $4\times 4\times 64$ (stride 2), Batch norm, ELU  \\
FC 256, Batch norm, ELU \\
FC 32 (Mean) $\Vert$ FC 32, Sigmoid (Std. dev.)  \\
 \bottomrule 
\end{tabular}
\vspace{-1mm}
\caption{Architectural details of the VAE model used for CelebA and SVHN datasets. The architecture of the generator is reverse of the encoder with Conv layers replaced with transposed convolutions (Deconv) and an additional final Deconv layer of size $4\times 4\times 3$.}
\label{tab:arch}
\end{table}

\subsection{Synthetic Manifold}
Since it is difficult to visualize high dimensional real data as manifolds, we illustrate the geodesic traversal using a simple analytically defined manifold. In particular, we use a hyperbolic paraboloid which is a 2-D surface in three dimensions, defined as the set $\{(a,b,c): a=z_1, b=z_2, c=(z_1^2-z_2^2), (z_1,z_2)\in \reals^2\}$. We sample data from this manifold using ancestral sampling, with $z_1,z_2\sim N(0,1)$ and $c=(z_1^2-z_2^2)$. We sample $50k$ points on this manifold and train a VAE on this data with latent dimension of $2$. The encoder $h$ is a two layer neural network with the fully-connected hidden layer of size $100$ (FC-100) having ELU activations. 
The encoder outputs the mean (FC-2) and variance (FC-2, followed by Sigmoid) of the approximate posterior. The decoder $g$ has reverse architecture of the encoder (FC-100, ELU, FC-3) and maps the two dimensional latents to three dimensional points on the manifold. We use exponential linear units (ELU) \cite{clevert2015fast} so that the resulting generator mapping is differentiable ($C^1$). Although the use of ELUs does not result in a $C^\infty$ mapping, it does ensure that we generate a $C^1$ manifold. Also, all of our proposed algorithms are valid because they require at most first derivatives of the generator. 
We train this using minibatch stochastic gradient descent with batch size of $100$ and learning rate of $0.0001$ for $100k$ minibatch iterations. 

We pick two points reasonably far away on the analytically defined hyperbolic paraboloid, $x=(-3,-3,0)$ and $y=(3,-3,0)$, and map these to the latent space of the trained VAE using the encoder as $z_x=h(x)$ and $z_y=h(y)$, where $h$ in this context represents the mean of the approximate posterior.
The corresponding points on the manifold are obtained as $g(z_x)$ and $g(z_y)$. We use Algorithm \ref{alg:geodesic} to estimate the geodesic connecting the points $g(z_x)$ and $g(z_y)$, and compare it with the curve traced on the generator's manifold by linear interpolation between $z_x$ and $z_y$. 

Fig.~\ref{fig:synthgeo} visualizes the true shape of our analytically defined hyperbolic paraboloid (left-most plot) along with the shape of the manifold as learned by the VAE's generator (middle plot). We also visualize the geodesic and linear interpolation curves between the points $g(z_x)$ and $g(z_y)$ on the learned manifold (middle plot), and the same set of curves between $z_x$ and $z_y$ in the two dimensional latent space (right-most plot). This clearly brings out the differences between linear and geodesic interpolation paths, with a shorter geodesic curve on the manifold (about 35\% smaller arclength than the linear curve) being traced by a longer curve in the latent space.


\subsection{Real Manifolds}
In this section, we investigate the Riemannian geometry of the generated manifolds learned on real images by carrying out computations such as geodesic interpolation and geodesic mean, and comparing these with the corresponding linear counterparts in $Z$ space. 
We use two real image datasets in our experiments: \\
{\bf CelebA}\cite{liu2015deep}.~ It consists of $202,599$ RGB face images of celebrities. We use  center-cropped images of shape $64\times 64\times 3$ as used in several earlier works, using $80\%$ of these for training the VAE. \\
{\bf SVHN} \cite{netzer2011reading}.~ It consists of house numbers obtained from Google Street View images. We use about $530k$ cropped digits of shape $32\times 32\times 3$ for training that are provided as part of the dataset. 

\noindent {\bf Implementation details.~} Architecture of the encoder ($h:X\to Z$) for both CelebA and SVHN is shown in  Table~\ref{tab:arch}. The architecture for the generator ($g:Z\to X$) is reverse of the encoder architecture with an additional transposed convolution layer that outputs the RGB image. The latent dimension is kept at $32$ for both datasets. 
 The model is trained for $50k$ minibatch iterations (batch size of $100$) using ADAM \cite{kingma2014adam} with the learning rate of $0.0002$.  

\begin{figure}
\begin{tabular}[t]{cc}
    65.84& \adjustimage{height=1cm,valign=m}{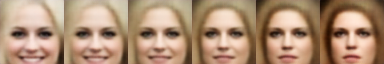} \\
    65.17 & \adjustimage{height=1cm,valign=m}{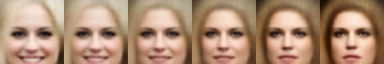} \\
    56.34 & \adjustimage{height=1cm,valign=m}{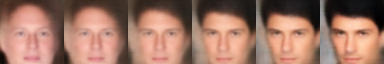} \\
    53.76 & \adjustimage{height=1cm,valign=m}{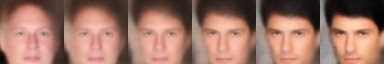} \\
    82.71 & \adjustimage{height=1cm,valign=m}{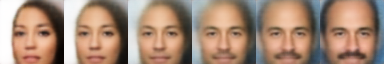} \\
    77.01& \adjustimage{height=1cm,valign=m}{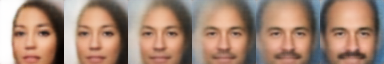} \\
\end{tabular}
\caption{Linear and geodesic interpolation results for CelebA dataset. Rows 1, 3, 5: linear; Rows 2, 4, 6: geodesic; Column 1: arc length.}
\label{tbl:GeoderpC}
\vspace{-4mm}
\end{figure}

\subsubsection{Geodesic Interpolation}
\begin{figure}
\begin{tabular}[t]{cc}
    18.10& \adjustimage{height=1cm,valign=m}{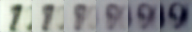} \\
    17.88 & \adjustimage{height=1cm,valign=m}{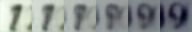} \\
    24.03 & \adjustimage{height=1cm,valign=m}{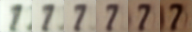} \\
    23.88 & \adjustimage{height=1cm,valign=m}{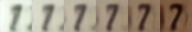} \\
    15.94 & \adjustimage{height=1cm,valign=m}{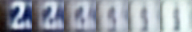} \\
    15.66& \adjustimage{height=1cm,valign=m}{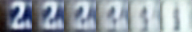} \\
\end{tabular}
\caption{Linear and geodesic interpolation results for SVHN dataset. Rows 1, 3, 5: linear; Rows 2, 4, 6: geodesic; Column 1: arc length.}
\label{tbl:GeoderpS}
\vspace{-5mm}
\end{figure}

We use Algorithm \ref{alg:geodesic} to estimate the geodesic curve connecting a given pair of images on the generated manifold, discretizing it at 10 points ($T=10$). To get an image on the generated manifold, we pick a real image $x$ from the dataset and use $g(h(x))$ to get the corresponding point on the generated manifold.   Fig~\ref{tbl:GeoderpC} and \ref{tbl:GeoderpS} show a few images (equally spaced in Z space) on the linear and geodesic interpolation curves along with their arclengths, for CelebA and SVHN, respectively. 
Although, the geodesic curve on the manifold gives a shorter arclength than linear interpolation in Z space, 
the difference is not as pronounced as observed in our earlier experiment with synthetic manifold. 
This suggests that the generated manifolds learned by our VAE architecture for CelebA and SVHN, although nonlinear, have very little curvature.

\subsubsection{Fr\'echet Means}
We take a step further and look at the Fr\'echet mean of a chosen set of points on the generated manifold, comparing it with the linear mean in Z space. 
The Fr\'echet mean of a set is a point on the manifold which minimizes the total sum-of-squared geodesic distance to all the points in the set. In our setting, if $z_1, \ldots, z_N \in Z$ are input data points, the Fr\'echet mean is defined as the solution to the optimization problem:
$$\hat{\mu} = \arg \min_{\mu \in Z} \sum_{i=1}^N d(\mu, z_i)^2,$$
where $d(\mu, z_i)$ is the geodesic distance, i.e., the arc length of path computed using Algorithm \ref{alg:geodesic}. We optimize this least squares problem using gradient descent in the latent coordinates for $\mu$.

\begin{figure}[t]
    \centering
    \begin{subfigure}[b]{0.1\textwidth}
        \includegraphics[width=\textwidth]{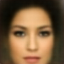}
    \end{subfigure}
    \begin{subfigure}[b]{0.1\textwidth}
        \includegraphics[width=\textwidth]{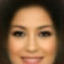}
    \end{subfigure}
    \begin{subfigure}[b]{0.1\textwidth}
        \includegraphics[width=\textwidth]{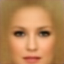}
    \end{subfigure}
    \begin{subfigure}[b]{0.1\textwidth}
        \includegraphics[width=\textwidth]{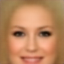}
    \end{subfigure}
    
     \begin{subfigure}[b]{0.1\textwidth}
        \includegraphics[width=\textwidth]{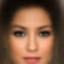}
    \end{subfigure}
    \begin{subfigure}[b]{0.1\textwidth}
        \includegraphics[width=\textwidth]{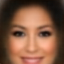}
    \end{subfigure}
    \begin{subfigure}[b]{0.1\textwidth}
        \includegraphics[width=\textwidth]{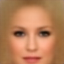}
    \end{subfigure}
    \begin{subfigure}[b]{0.1\textwidth}
        \includegraphics[width=\textwidth]{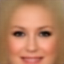}
    \end{subfigure}
    \caption{Linear mean (top) and Geodesic mean (bottom) in $Z$ space for the four groups of images from CelebA. From left to right: \emph{(black hair, mouth close)}, \emph{(black hair, mouth open)}, \emph{(blond hair, mouth close)}, \emph{(blond hair, mouth open)}.}\label{fig:mean}
    \vspace{-4mm}
\end{figure}

A set of real images from CelebA is constructed by randomly selecting images from the dataset that all have the same value for a chosen pair of attributes.  We construct four such sets, each consisting of $100$ images, corresponding to attributes \emph{(black hair, mouth open)}, \emph{(black hair, mouth closed)}, \emph{(blond hair, mouth open)} and \emph{(blond hair, mouth closed)}, respectively. 
We find the corresponding points on the VAE's generated manifold by applying function $g\circ h$ on each of these images. 
Fig.~\ref{fig:mean}  visualizes the Fr\'echet means and linear means for these four groups of images. Here the Fr\'echet means are similar in appearance to the linear means in the latent space. Again, this indicates that there may be limited curvature in the manifold. However, there are certainly subtle differences (particularly in the color) that indicates curvature is playing at least some role.


\subsubsection{Geodesic Distance and Attribute Groupings}
In this section, we analyze how well are the geodesic distances aligned with the groupings of the images based on the ground truth attributes. We reuse the four groups of images constructed in the earlier section for CelebA for this experiment. In addition, we also construct ten groups of images for SVHN, with each group consisting of $50$ randomly sampled images of a digit. We apply $g\circ h$ on each of these points to get corresponding points on the generated manifold, and compute linear and geodesic distances for each pair of these points. This gives us linear and geodesic distance matrices of size $400\times 400$ for CelebA and $500\times 500$ for SVHN. 
We calculate $R^2$ scores for each distance matrix, $R^2 = 1-\frac{ \sum_{(ij:l_i=l_j)} d^2(x_i,x_j)}{\sum_{i,j=1}^N d^2(x_i,x_j)}$, where $l_i$ is the attribute label for $x_i$ and $N$ is just total number of data points. The $R^2$ score essentially measures the ratio of the intra-group squared distances and total squared distances, with a higher $R^2$ value indicating better agreement between the attribute based grouping and the distances. As $R^2$ score is already normalized by the sum of all squared distances, it is directly comparable across linear and geodesic distance matrices. As shown in Table~\ref{table:cluster}, we obtain slightly higher $R^2$ scores with geodesic distances compared to the linear distances, indicating that geodesic distances group similar images slightly closer together than linear distances.

\begin{figure*}[t]
    \begin{center}
  \includegraphics[width=0.45\textwidth]{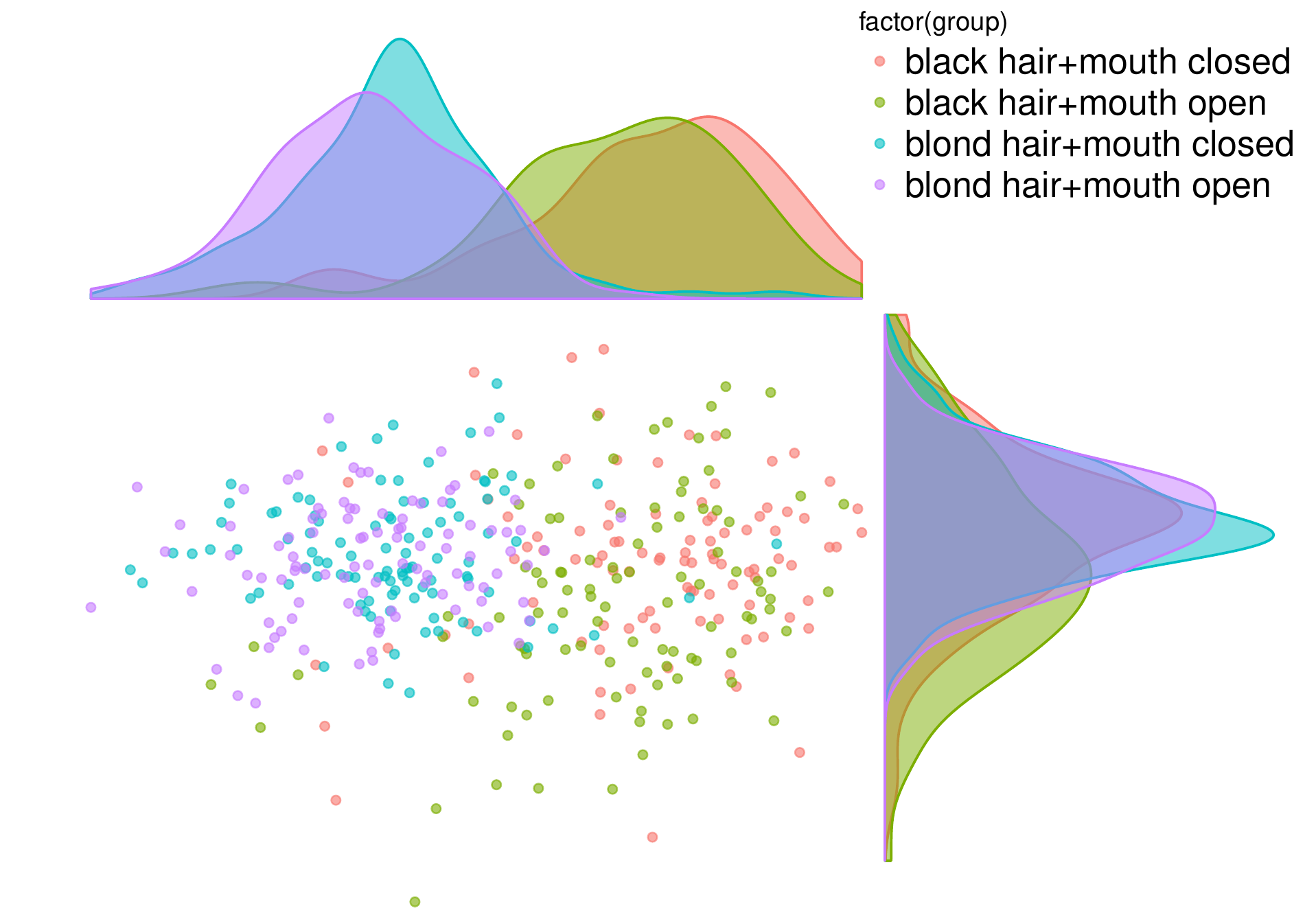}
  \includegraphics[width=0.45\textwidth]{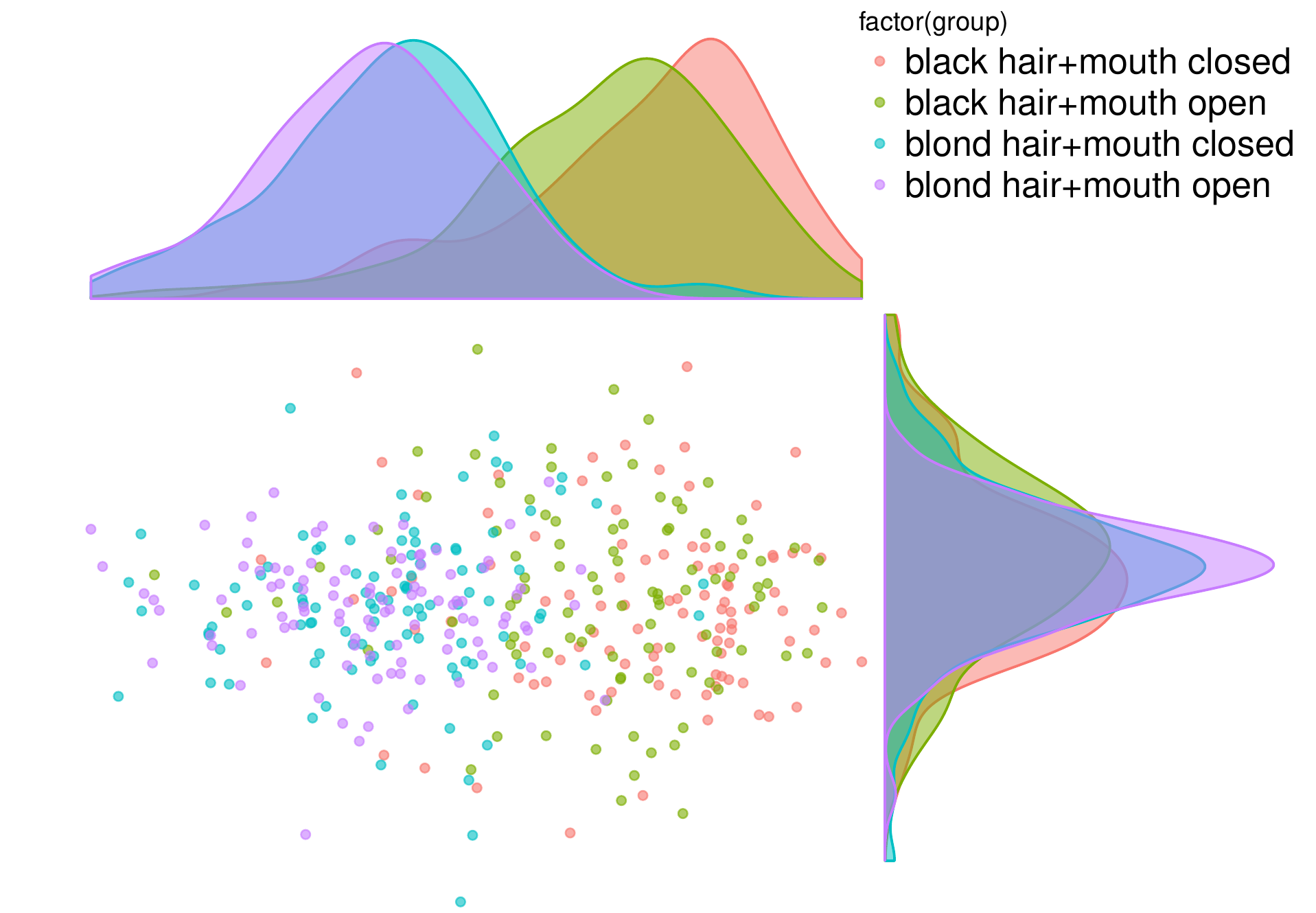}
\end{center}
\vspace{-6mm}
\caption{MDS embedding for  linear (left) and geodesic (right) distance matrices, for four groups of images from CelebA.}
\label{fig:geodistance}
\vspace{-4mm}
\end{figure*}

\begin{figure*}[t]
    \begin{center}
  \includegraphics[width=0.48\textwidth]{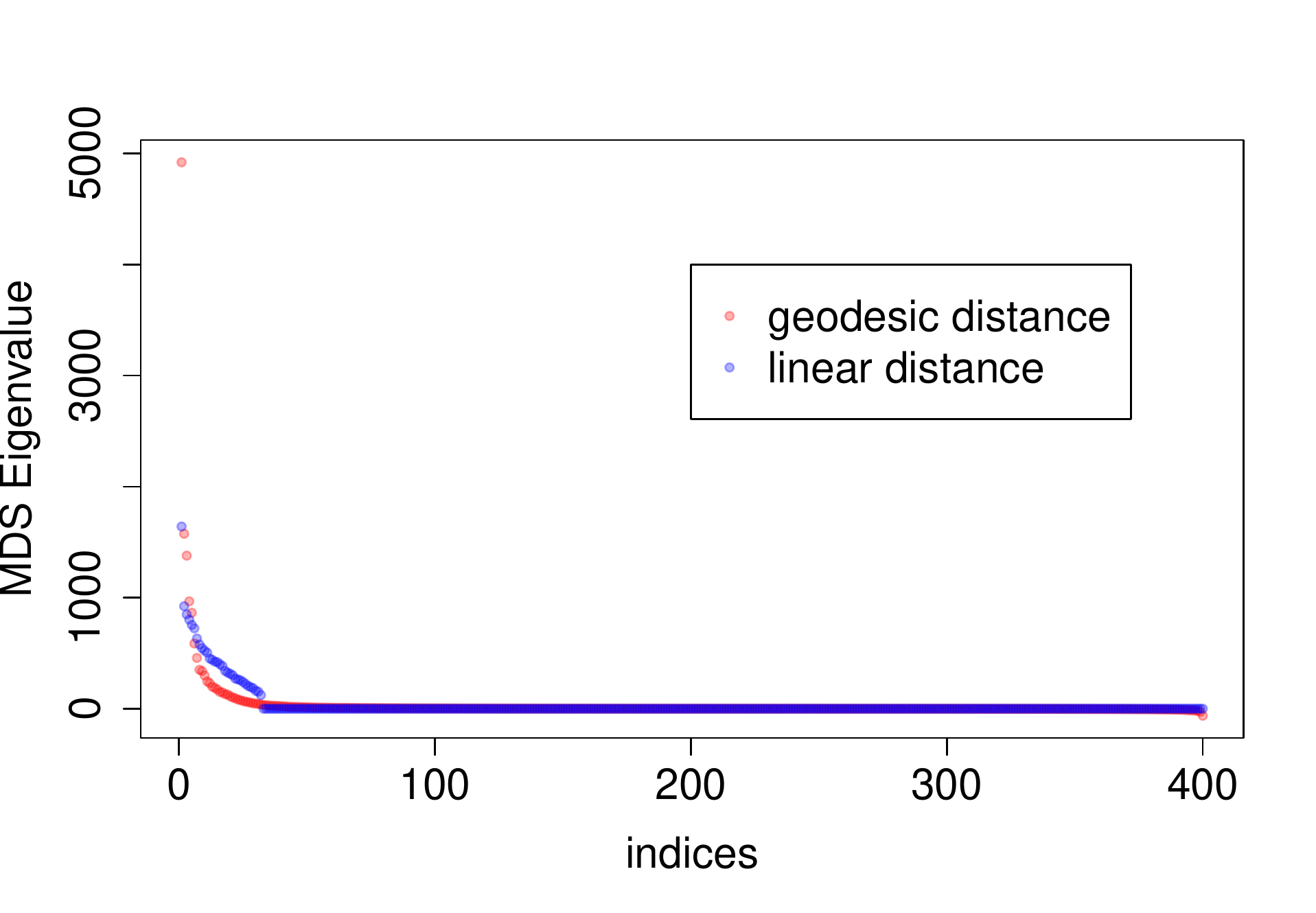}
  \includegraphics[width=0.48\textwidth]{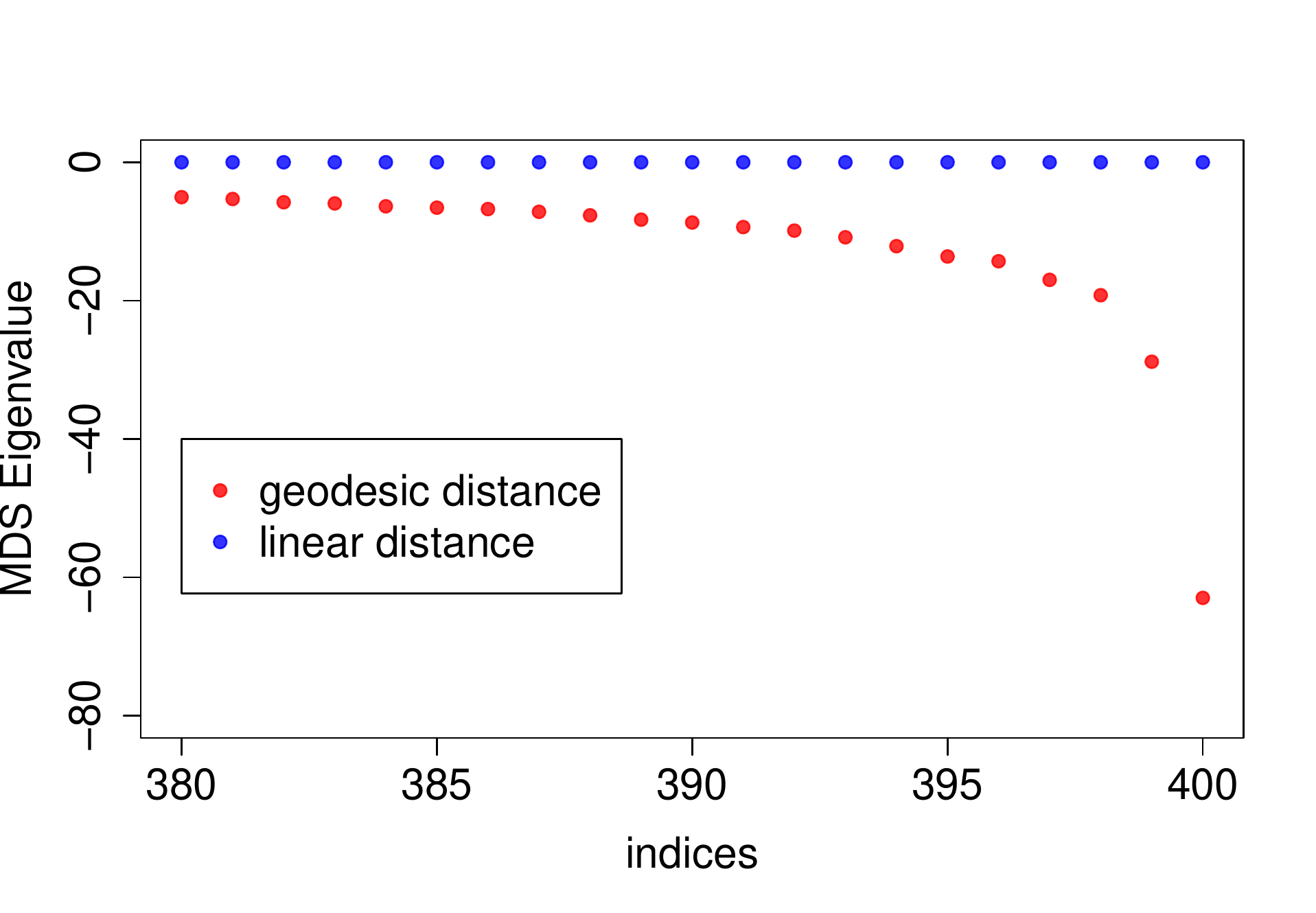}
\end{center}
\vspace{-7mm}
\caption{Eigenvalues of the MDS matrices for the four groups of images from CelebA dataset ($400$ total images). Left: all 400 eigenvalues. Right: Zooming in on lowest eigenvalues. Note that the vertical axis scale is much smaller in the right plot.}
\label{fig:MDS}
\vspace{-2mm}
\end{figure*}

We also use multidimensional scaling (MDS) to embed the points into two dimensions based on these distance matrices, which are visualized in Fig.~\ref{fig:geodistance} for CelebA. The embedding based on geodesic distances visually seem to give a slightly tighter concentration around the groups, compared to the embedding based on linear distances. We also calculate the eigenvalues for the MDS matrices and plot them in Fig. \ref{fig:MDS}. The eigenvalues of MDS explain whether the data can be isometrically embedded in Euclidean space (i.e., while preserving the distance metric between pairs of points). If all eigenvalues are non-negative, then this Euclidean embedding is possible, and the dimension of the Euclidean space is the number of nonzero eigenvalues. The presence of negative eigenvalues demonstrate that the space has nonzero curvature, and exact Euclidean embedding is impossible. The magnitude of the negative eigenvalues is a measure of how far the manifold distances are deviating from Euclidean, i.e., it is a measure of how much curvature the manifold has. As expected, the linear distance matrix resulted in exactly $32$ positive eigenvalues, with exactly zero eigenvalues after $32$. The geodesic distance matrix has negative eigenvalues, but they have very small magnitude compared with the positive eigenvalues. This strongly indicates that the generated manifold has some curvature, but it is close to being zero.

\begin{table}[t]
\begin{center}
\begin{tabular}[t]{c|c|c}
	&Geodesic&Linear\\
    \hline
    CelebA&  \bf 0.7782278&0.7638913\\
    \hline
    SVHN & \bf 0.9024925&0.9021349 \\
\end{tabular}
\end{center}
\vspace{-3mm}
\caption{$R^2$ Scores with geodesic and linear distance matrices (the higher the better)}
\label{table:cluster}
\vspace{-3mm}
\end{table}


\subsubsection{Geodesic Analogy}
The analogy problem is defined as $a:b::c:?$. In our context, $a,b$ and $c$ are images and we want to find an image that is related to $c$ in the same way as $a$ is related to $b$. We reuse the four CelebA groups constructed in the earlier experiments. We take $a$ to be the geodesic mean of the group \emph{(blond hair, mouth closed)} and $b$ to be the geodesic mean of the group \emph{(blond hair, mouth open)}. We take $c$ to be a randomly selected test image with attributes \emph{(blond hair, mouth closed)}. For geodesic analogy, we first compute the geodesic between $a$ and $b$. The initial velocity vector at $a$ is then parallel translated to $c$ along the geodesic connecting $a$ and $c$ using Algorithm \ref{alg:parallel}. We then use Algorithm \ref{alg:shoot} to \emph{shoot} a geodesic of same arc length as the $a$-$b$ geodesic along this parallel translated vector. The end point of this geodesic is expected to have a similar semantic relation to $c$, as $b$ is related to $a$ (\ie, change in the binary \emph{mouth} attribute from \emph{close} to \emph{open}). 

\begin{figure}[t]
    \centering
    \begin{subfigure}[b]{0.45\textwidth}
        \includegraphics[width=\textwidth]{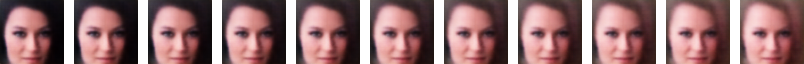}
    \end{subfigure}
    \begin{subfigure}[b]{0.45\textwidth}
        \includegraphics[width=\textwidth]{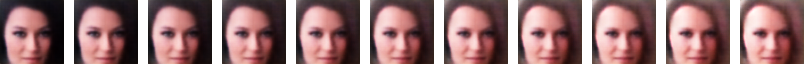}
    \end{subfigure}
    \begin{subfigure}[b]{0.45\textwidth}
        \includegraphics[width=\textwidth]{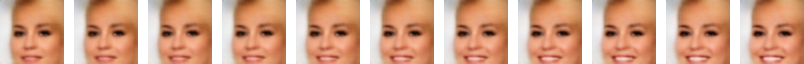}
    \end{subfigure}

    \begin{subfigure}[b]{0.45\textwidth}
        \includegraphics[width=\textwidth]{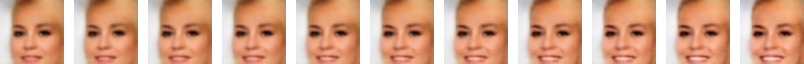}
    \end{subfigure}
    \vspace{-2mm}
    \caption{Linear mean vector analogy (rows 1,3) \emph{vs.} geodesic parallel translated vector analogy (rows 2,4). First two rows change black hair to blond, and the last two rows change closed mouth to open.}\label{fig:Cgeoderp}
    \vspace{-4mm}
\end{figure}

We also try a linear analogy operation in $Z$ space. We compute the difference $(z_b-z_a)$ (where $a=g(z_a)$, $b=g(z_b)$), and add the resulting vector to $z_c \in Z$ corresponding to the test image $c$ (\ie, $c = g(z_c)$). The answer to the linear analogy problem is then taken to be the image $y = g(z_b - z_a + z_c)$. 
Fig. \ref{fig:Cgeoderp} shows the results for geodesic and linear analogies for two different attribute combinations. The linear analogy is visually quite close to the geodesic analogy (with subtle differences), which again suggests that the generated manifold has very low curvature.

\section{Conclusion}
\vspace{-2mm}
In this paper we have introduced methods for exploring the Riemannian geometry of manifolds learned by deep generative models. Our experiments show that these models represent real image data with manifolds that have surprisingly little curvature. Consequently, straight lines in the latent space are relatively close to geodesic curves on the manifold. This fact may explain why traversal in the latent space results in visually plausible changes to the generated data: curvilinear distances in the original data metric are roughly preserved. However, our experiments were limited to a single type of deep network (VAE) and two real image data sets (CelebA and SVHN). Further investigation into this phenomenon is warranted, to identify if there are other architectures or datasets where curvature plays a more prominent role. Also, even for the results presented here, the role of curvature should not be completely discounted: there are still differences between latent distances and geodesic distances that may have more nuanced effects in certain applications. We believe that exploring the Riemannian geometry of deep generative models, using the tools developed in this paper, will be an important step in understanding the high-dimensional, nonlinear spaces these models learn. 

{\small
\bibliographystyle{ieee}
\bibliography{ml}
}

\end{document}